\pdfoutput=1

\documentclass[11pt]{article}

\usepackage[]{EMNLP2023}

\usepackage{times}
\usepackage{latexsym}

\usepackage[T1]{fontenc}

\usepackage[utf8]{inputenc}

\usepackage{microtype}
\usepackage{multirow} 
\usepackage{inconsolata}

\usepackage{tikz}
\usetikzlibrary{arrows.meta,shapes.geometric,positioning,fit,calc,decorations.pathreplacing}

\usepackage{listings}
\usepackage{amsmath}
\title{It Takes Two: A Dual Stage Approach for Terminology-Aware Translation}

\author{Akshat Singh Jaswal \\
  PES University\\
  \texttt{sja.akshat@gmail.com} \\}

\begin{document}
\maketitle
\begin{abstract}
This paper introduces DuTerm, a novel two-stage architecture for terminology-constrained machine translation. Our system combines a terminology-aware NMT model, adapted via fine-tuning on large-scale synthetic data, with a prompt-based LLM for post-editing. The LLM stage refines NMT output and enforces terminology adherence. We evaluate DuTerm on English-to-German, English-to-Spanish, and English-to-Russian for the WMT 2025 Terminology Shared Task. We demonstrate that flexible, context-driven terminology handling by the LLM consistently yields higher quality translations than strict constraint enforcement. Our results highlight a critical trade-off, revealing that an LLM's intrinsic knowledge often provides a stronger basis for high-quality translation than rigid, externally imposed constraints.
\end{abstract}

\section{Introduction}
The accurate and consistent translation of domain-specific terminology is a challenge in the field of Machine Translation and is of importance in domains such as law, medicine, and engineering, where precision is critical~\citep{NAVEEN2024110878}. While modern Neural Machine Translation systems based on architectures like the Transformer have achieved remarkable fluency and quality on general text, their performance in terminology-constrained texts remains a critical area for improvement~\citep{Vaswani2017, bahdanau2016neuralmachinetranslationjointly, johnson2017googlesmultilingualneuralmachine}. This issue is particularly relevant given findings of recent WMT shared tasks, which have consistently highlighted the need for systems that can effectively handle domain-specific vocabulary~\citep{Post2018}. The WMT 2025 Terminology Shared Task ~\citep{WMT2025} provides a focused platform to evaluate MT systems ability to handle domain-specific terminology under controlled conditions across multiple language pairs: English to German, English to Spanish, and English to Russian.

Previous research into terminology-constrained MT can be broadly categorized into two main approaches: inference-time methods and training-time methods. Inference-time approaches incorporate terminology constraints directly into the decoding process, often through techniques like constrained beam search or by re-ranking n-best lists of candidate translations~\citep{Zhang2023}. While these methods are highly effective at enforcing constraints, they can be computationally expensive and may compromise the overall fluency and grammatical correctness of the output by forcing the model to generate awkward or unnatural phrases. Recent work has explored ways to make these methods more efficient, but the trade-off between enforcing terminology constraints and fluency remains a key consideration~\citep{Moslem2023}.

Alternatively, training-time methods aim to teach models how to handle terminology constraints by integrating the terminology information into the training data itself. This is commonly done through the use of special tags that surround the terms to be translated~\citep{Dinu2019}. This approach allows the model to learn how to produce more natural and grammatically correct output, but it provides no guarantee that all constraints will be respected during inference~\citep{Susanto2020}.

We present DuTerm, a two-stage architecture that addresses these limitations by combining the strengths of both training-time and inference-time methodologies. We recognize that terminology-constrained translation is not merely a lexical substitution problem but requires a deeper understanding of linguistic context, especially when dealing with the complex morphology of languages like German and Russian. Our system is specifically designed to tackle the multifaceted evaluation framework of the WMT 2025 shared task.

\section{Method}

\subsection{Terminology-Aware Neural Machine Translation}
\paragraph{Overview}
We develop a terminology aware MT model via large-scale, tagged synthetic data and targeted fine-tuning. The pipeline: extract and analyze terminology, generate tagged, context-rich parallel data (single-term and multi-term) and standardize tags and ensure annotation consistency. We also quality-filter with COMET$_{\text{QE}}$ \citep{rei-etal-2022-cometkiwi} and deduplicate , this finally adapts a multilingual NMT model with parameter efficient fine-tuning .

\paragraph{Terminology Extraction and Analysis}
We parse the WMT 2025 dev files for English$\rightarrow$German/Spanish/Russian to build bilingual terminology dictionaries. The dictionaries typically exceed 1,000 unique pairs per direction. We track terms and occurrences using \texttt{repetition\_id}s. We also use the LLM to generate more terms similar to the terms provided in the dictionaries.

\paragraph{Synthetic Data Generation}
We use GPT-4o \citep{openai2024gpt4ocard} to create parallel sentences that naturally embed required terms and explicitly insert boundary tags (\texttt{[TERM]}...\texttt{[/TERM]}) on both source and target. There are two modes we use to generate these parallel sentences

\emph{Single-term mode}: generates sentence pairs containing exactly one term instance per sentence.

\emph{Multi-term mode}: randomly selects 2–3 term pairs to appear together, teaching co-occurrence handling and disambiguation. 

We employ temperature sampling (0.3–0.7), concurrent generation, and strict parsing to yield well-formed bilingual pairs.

\paragraph{Tag Standardization and Quality Filtering}
A re-tagging pass enforces consistent annotation, longest-first matching prevents partial shadowing, case-insensitive detection with original case preservation, and inverse mapping ensures symmetric target-side tagging.
Each pair is scored by COMET$_{\text{QE}}$. We deduplicate on the source side and keep only high-confidence items using a conservative threshold (0.85-0.9) depending on the language, typically retaining 60–70\% of outputs, yielding $\sim$10k–15k pairs per language direction.

\paragraph{Multilingual Model Adaptation}
For the foundation translation model, we select NLLB-200 3.3B, a state-of-the-art multilingual neural machine translation model with demonstrated strong performance across our target languages \citep{nllbteam2022languageleftbehindscaling}. This model provides robust baseline capabilities while supporting the specialized terminology handling adaptations we require.

We extend the model's vocabulary with terminology markup tokens to ensure atomic treatment of terminology annotations. This prevents subword tokenization from fragmenting our special markup, ensuring that terminology boundaries are consistently preserved during training and inference.

The training process employs several optimization strategies designed for stable, effective adaptation. The process also combines filtered datasets from all three target languages, creating unified multilingual adaptation that benefits from cross-lingual transfer.

\subsection{LLM-Based Post-Editing }
\paragraph{Overview}
An LLM refines the NMT output given the source sentence and required term pairs, enforcing strict terminology adherence while improving fluency and morphology.\citep{raunak2023leveraginggpt4automatictranslation}

\paragraph{Post-Editing Procedure}
We use prompts that present the source, translation, and provide explicit source to target term mappings. The LLM is instructed to preserve meaning, apply the exact target terms, maintain tags where required, and improve readability without paraphrasing away constraints. The LLM we choose to use is GPT-4o \citep{openai2024gpt4ocard} due to its combination of high translation quality and relatively lower price.

\paragraph{Terminology-Aware Processing}
\emph{Dynamic resolution}: per-input selection of proper/random/no-term constraints from reference terminology databases with whitespace-normalized matching.  
\emph{Mode-adaptive behavior}: when constraints exist, the LLM must enforce them; otherwise it performs quality-only edits while being sensitive to technical terms.  
\emph{Constraint satisfaction}: explicit mappings and formatting rules are included in the prompt; outputs must preserve required terminology and markup.

\paragraph{Quality Assurance and Robustness}
We run the LLM at low temperatures (0.3) for deterministic edits. Each hypothesis is validated for format, tag integrity, and constraint satisfaction before acceptance with a pre-existing parser.
We verify filename schemas, presence of all terminology modes per language pair, and JSONL structure. We assess quality with COMET$_{\text{QE}}$ (after tag stripping) and compute terminology preservation via exact-match checks on required terms. This ensures reliability of final outputs.

\section{Results}

\begin{table*}[t]
\centering
\normalsize
\begin{tabular}{lccccc}
\hline
\textbf{Lang} & \textbf{Type} & \textbf{BLEU} & \textbf{chrF2++} & \textbf{Prop. SR} & \textbf{Rand. SR} \\
\hline
\multirow{3}{*}{DE} 
 & noterm & 38.24 & 62.61 & 0.43 & 0.69 \\
 & proper & 48.06 & 70.74 & 0.98 & 0.73 \\
 & random & 43.77 & 67.22 & 0.48 & 0.99 \\
\hline
\multirow{3}{*}{ES} 
 & noterm & 45.98 & 67.05 & 0.47 & 0.73 \\
 & proper & 58.51 & 76.08 & 0.99 & 0.78 \\
 & random & 53.28 & 72.05 & 0.49 & 0.98 \\
\hline
\multirow{3}{*}{RU} 
 & noterm & 27.88 & 55.29 & 0.39 & 0.69 \\
 & proper & 35.80 & 63.57 & 0.98 & 0.72 \\
 & random & 32.25 & 59.85 & 0.42 & 0.99 \\
\hline
\end{tabular}
\caption{\label{tab:results} 
Evaluation results for English$\rightarrow$German (DE), English$\rightarrow$Spanish (ES), and English$\rightarrow$Russian (RU) across three terminology handling strategies. Metrics include BLEU, chrF2++, and terminology success rates (proper and random).
}
\end{table*}

We evaluate the system using three complementary metrics used by the WMT organizers: BLEU for overall translation adequacy, chrF2++ for character-level fluency and robustness, and terminology success rates (proper and random) to directly measure constraint satisfaction \citep{papineni-etal-2002-bleu,popovic-2015-chrf}. Results are reported for English$\rightarrow$German (DE), English$\rightarrow$Spanish (ES), and English$\rightarrow$Russian (RU) across three terminology strategies: \textit{noterm}, \textit{proper}, and \textit{random}.

Table~\ref{tab:results} summarizes the findings. Several clear patterns emerge:

1. \textbf{Strict terminology enforcement (\textit{proper})} achieves the highest BLEU and chrF2++ across all languages (48.06 for DE, 58.51 for ES, 35.80 for RU), indicating improved lexical precision and sentence-level quality when constraints are respected. It also yields near-perfect proper terminology success rates ($\geq$0.97).

2. \textbf{Unconstrained translation (\textit{noterm})} consistently underperforms, producing the lowest BLEU and chrF2++ values across languages (e.g., 38.24 BLEU in DE and 27.88 in RU). While fluency remains reasonable, failure to enforce constraints leads to poor terminology precision.

3. \textbf{Random terminology enforcement} produces intermediate BLEU/chrF2++ but near-perfect random-term success rates ($\sim$0.98). This highlights that while the model can force arbitrary terminology, doing so compromises contextual appropriateness.

4. \textbf{Language-specific trends} align with expectations: Spanish shows the highest overall scores, reflecting its structural similarity to English. Russian shows the widest gap between \textit{proper} and \textit{noterm}, emphasizing the difficulty of morphology-rich languages for terminology control.

Overall, these results demonstrate that while strict enforcement maximizes terminology accuracy and boosts surface-level quality metrics, it can occasionally reduce flexibility. In contrast, unconstrained approaches produce more natural translations but risk terminology inconsistency.

\section{Conclusion}
This paper presents our approach to the terminology shared task, focusing on English to German, Spanish, and Russian translation directions. Our system leverages LLMs to improve existing translations with varying terminology handling strategies. Our results demonstrate that allowing the LLM to flexibly handle terminology often yields higher translation quality than strict terminology enforcement. These findings highlight the potential of prompt-based LLM systems for technical and business translation tasks, and provide insights into effective strategies for terminology management in neural translation workflows. The intution behind why this approach works so well is that NMTs often excel at strict word-level translations however they can struggle with context-dependent nuances. Our approach leverages post-editing with a LLM on top of the initial NMT outputs. By starting from a reliable NMT translation, the post-editing model receives a structured, partially correct target sentence, which allows it to focus on higher-level improvements resolving ambiguities, adjusting word order, and refining context. This guided refinement is often more effective than translating directly from scratch with an LLM or NMT which must simultaneously handle term accuracy and contextual fluency.

For future work, exploring adaptive learning mechanisms that integrate terminology dynamically, rather than relying on static prompts, could enhance robustness across domains and languages. End-to-end or memory-augmented architectures that maintain consistency across sentences and documents hold promise for more coherent outputs. Expanding evaluations to other language models and diverse, domain-specific corpora would help validate the approach’s generalizability and reveal domain-dependent challenges. Incorporating hybrid strategies, such as combining prompt guidance with fine-tuning or reinforcement learning, and enabling user-driven interaction for terminology control could further improve usability and accuracy. Together, these directions offer a pathway toward more flexible, context-aware, and widely applicable terminology-aware translation systems.

\section*{Limitations}

Our approach, based on a prompt-driven framework, faces several limitations. It depends heavily on carefully crafted prompts, which may not generalize well across domains, languages, or model architectures. The sequential processing of terminology matching and translation refinement limits the system’s ability to adaptively enforce terminology constraints. Furthermore, operating at the sentence level overlooks opportunities for document-level consistency and context-aware terminology usage, which are crucial in practical translation tasks. Our evaluation, conducted solely on GPT-4o \citep{openai2024gpt4ocard}, restricts the generalizability of findings, and focusing on technical and business domains may not capture challenges present in specialized fields like medical or legal translation. Additionally, while COMET$_{\text{QE}}$ ,BLEU ,chrF++ provide automated scalability, it may not fully reflect terminological precision and contextual appropriateness, suggesting the need for complementary evaluation methods that include human judgment.

\bibliography{anthology,custom}
\bibliographystyle{acl_natbib}

\appendix
\newpage
\onecolumn
\section{Prompts}
\label{sec:appendix}

We include below the full prompts used in our experiments for reproducibility.  

\subsection{Single-Term Prompt}
\begin{lstlisting}[caption={Prompt template for generating bilingual sentence pairs with a single terminology constraint.}]
Generate {n} professional, domain-specific English-({target_lang}) bilingual sentence pairs for terminology translation.
The term pair to use is: {source_term}\(EN) : \"{target_term}\ ({target_lang})
Requirements:
- Each sentence pair must be natural, fluent, and contextually appropriate for IT or financial domains.
- Include the term exactly once per sentence.
- Wrap the term with [TERM] and [/TERM] in both the English and ({target_lang}) sentences.
- Ensure accurate translation and alignment of the term.
Format:
EN: [sentence]
{target_lang}: [sentence]
Output exactly {n} such pairs.
\end{lstlisting}

\subsection{Multi-Term Prompt}
\begin{lstlisting}[caption={Prompt template for generating bilingual sentence pairs with multiple terminology constraints.}]
Generate {n} professional, domain-specific English-({target_lang}) bilingual sentence pairs for terminology translation.
Use ALL of the following term pairs in each sentence pair:\n{terms_str}
Requirements:
- Each sentence pair must be natural, fluent, and contextually appropriate for IT or financial domains.\n"
- Include each term exactly once per sentence.
- Wrap each term with [TERM] and [/TERM] in both the English and ({target_lang}) sentences.\n"
- Ensure accurate translation and alignment of the terms.
Format:
EN: [sentence]
{target_lang}: [sentence]
Output exactly {n} such pairs.
\end{lstlisting}

\subsection{Post-Editing with Terminology}
\begin{lstlisting}[caption={Prompt for post-editing with explicit terminology mappings.}]
As an expert English-{target_lang} translator specializing in technical and business documentation, improve this {target_lang} translation.

SOURCE (English): {source}

CURRENT TRANSLATION ({target_lang}): {translation}

REQUIRED TERMINOLOGY (English: {target_lang}): {term_str}

YOUR TASK:
1. Ensure all required terminology is correctly used
2. Maintain the same meaning as the source text
3. Ensure natural, fluent {target_lang} that sounds like native content
4. Preserve formatting, numbers, and special characters
5. Match the tone and register of the original text

Return ONLY the improved {target_lang} translation with no explanations, notes, or other text.
\end{lstlisting}
\newpage
\subsection{Post-Editing without Terminology}
\begin{lstlisting}[caption={Prompt for post-editing without explicit terminology guidance.}]
As an expert English-{target_lang} translator specializing in technical and business documentation, improve this {target_lang} translation.

SOURCE (English): {source}

CURRENT TRANSLATION ({target_lang}): {translation}

Note: There may be important terminology in the source text that should be translated precisely and consistently. Please ensure any technical or business terms are rendered correctly in {target_lang}.

YOUR TASK:
1. Enhance the translation for fluency and accuracy
2. Maintain the same meaning as the source text
3. Ensure natural, fluent {target_lang} that sounds like native content
4. Preserve formatting, numbers, and special characters
5. Match the tone and register of the original text

Return ONLY the improved {target_lang} translation with no explanations, notes, or other text.
\end{lstlisting}

\end{document}